\documentclass[lettersize, journal]{IEEEtran}
\usepackage{setspace}
\usepackage{romannum}
\usepackage{caption}
\usepackage{subfigure}
\usepackage{bm}
\usepackage{upgreek}
\usepackage{amsmath,amsfonts}
\usepackage{algorithmic}
\usepackage{algorithm}
\usepackage{array}
\usepackage{textcomp}
\usepackage{stfloats}
\usepackage{url}
\usepackage{verbatim}
\usepackage{graphicx}
\usepackage{cite}
\usepackage{xcolor}
\begin{document}
	
	\title{Resource Allocation for Twin Maintenance and Task Processing in Vehicular Edge Computing Network}
	
	\author{Yu Xie, Qiong Wu,~\IEEEmembership{Senior Member,~IEEE,} Pingyi Fan,~\IEEEmembership{Senior Member,~IEEE,} \\
	Nan Cheng,~\IEEEmembership{Senior Member,~IEEE,}
	Wen Chen,~\IEEEmembership{Senior Member,~IEEE,} \\
	Jiangzhou Wang,~\IEEEmembership{Fellow,~IEEE} and Khaled B. Letaief, ~\IEEEmembership{Fellow,~IEEE}
		\thanks{This work was supported in part by Jiangxi Province Science and Technology Development Programme under Grant 20242BCC32016; in part by the National Natural Science Foundation of China under Grant 61701197; in part by the National Key Research and Development Program of China under Grant 2021YFA1000500(4); in part by Shanghai Kewei under Grants 22JC1404000 and 24DP1500500; in part by the Research Grants Council under the Areas of Excellence Scheme under Grant AoE/E-601/22-R; and in part by the 111 Project under Grant B23008. (Corresponding author: Qiong Wu.)}
		\thanks{Yu Xie and Qiong Wu are with the School of Internet of Things Engineering, Jiangnan University, Wuxi 214122, China, and also with the School of Information Engineering, Jiangxi Provincial Key Laboratory of Advanced Signal Processing and Intelligent Communications, Nanchang University, Nanchang 330031, China (e-mail:  yuxie@stu.jiangnan.edu.cn, qiongwu@jiangnan.edu.cn).}
		\thanks{Pingyi Fan is with the Department of Electronic Engineering, State Key Laboratory of Space Network and Communications, and the Beijing National Research Center for Information Science and Technology, Tsinghua University, Beijing 100084, China (e-mail: fpy@tsinghua.edu.cn).}
		\thanks{Nan Cheng is with the State Key Lab. of ISN and School of Telecommunications Engineering, Xidian University, Xi’an 710071, China (e-mail: dr.nan.cheng@ieee.org).}
		\thanks{Wen Chen is with the Department of Electronic Engineering, Shanghai JiaoTong University, Shanghai 200240, China (e-mail: wenchen@sjtu.edu.cn).}
		\thanks{Jiangzhou Wang is with the School of Information Science and Engineering, Southeast University, Nanjing 211111, China. (email: j.z.wang@seu.edu.cn).}
		\thanks{Khaled B. Letaief is with the Department of Electrical and Computer Engineering, Hong Kong University of Science and Technology, Hong Kong (e-mail: eekhaled@ust.hk).}}
	
	
	
	\maketitle

	\begin{abstract}
		In the digital twin mobile edge network, the maintenance of the vehicle twin model and vehicular task processing in the server require the support of computing resources. In addition, they are performed simultaneously. Therefore, how to allocate resources for twin maintenance and task processing under limited server resources is crucial. However, current research tends to ignore the aspect of resource competition for twin maintenance. In this study, we analyze the delays of these two affected by resource allocation under a generic digital twin mobile edge network (DTMEN) to construct the optimization problem. For this problem, we transformed the problem using a Markov decision process. Meanwhile, we propose a multi-agent reinforcement learning (MADRL) based twin maintenance and task processing resource collaborative scheduling (TMTPRCS) algorithm to solve the problem. Experiments show that our proposed approach is effective in terms of resource allocation compared to other alternative algorithms.
		
	\end{abstract}
	
	
	\section{Introduction}
	\IEEEPARstart{W}{ith} the rapid advancement of fifth generation (5G) technology, there is a growing presence of vehicular applications and multimedia services in areas such as autonomous driving, navigation, high-definition video, and more. These advancements aim to enhance the overall driving experience [1]-[8] but also lead to an increase in the number of vehicular computing tasks that need to be addressed. However, the storage capacity and computing capabilities of vehicles are often insufficient to handle such demands, posing a challenge for real-time processing of computationally intensive tasks [9]-[15]. In response to this challenge, vehicle edge computing has emerged as a promising solution. By deploying a VEC server at roadside locations, it becomes possible to offload computational tasks from vehicles to these servers for processing and then return the results efficiently [16]-[25].
	
	Although VEC can provide computing services for vehicles, it also faces issues such as vehicle mobility and environmental dynamics during implementation[26]. Digital twin (DT), as a promising and emerging innovation, facilitates the creation of virtual models that accurately represent physical objects [27]-[29]. With advancements in 5G, edge computing, artificial intelligence, and related technologies, the capabilities of DT are continually enhancing. It now goes beyond one-way mirror simulation to enable two-way information interaction [30]. This bidirectional mapping offers a wealth of information for VEC network, enabling feature extraction and prediction of physical vehicles and their surrounding environment [31],[32]. Specifically, vehicles can transmit their data to the server using Vehicle-to-Everything (V2X) technology, allowing them to establish digital replicas on the server based on historical data, leading to the creation of DT models [33].
	
	The combination of DT and VEC can not only collect real-time operational data of vehicles in the VEC network, but also carry out the real time control and change of the status of vehicles[34]. Real-time vehicle information can be accessed and compared with historical data to identify and address potential problems or risks in vehicle operation proactively[35]. By deploying the DT at the VEC server, the DT can continuously monitor the physical state of the vehicle and network environment. This allows proactive decision making based on the latest data. In addition, DT acts as a virtual representation of the physical system, allowing decisions to be tested and validated in a risk-free environment, as opposed to deploying DRL directly to the server. This enhances the robustness of the decision-making process. However, this integration also presents challenges, particularly in terms of resource management. Current research under DTMEN tends to ignore the fact that digital twin maintenance can place demands on computing resources. Under DTMEN, vehicle digital twins are often built in VEC server. In addition, the VEC server will likewise need to allocate resources for vehicular computing tasks. And they are often performed simultaneously. Therefore, it becomes critical to allocate resources to meet the needs of both.
	
	In this paper, we consider a DT scenario for mobile edge network with multiple vehicles associated with a single VEC server and propose a multi-agent based twin maintenance and computing task processing resource collaborative scheduling algorithm (TMRCS)\footnote{The source code has been released at: https://github.com/qiongwu86/Resource-allocation-for-twin-maintenance-and-computing-tasks-in-digital-twin-mobile-edge-network.}, which maximizes the resource utility of computing resource allocation while ensuring the time limit requirements of twin maintenance and vehicle computing tasks. The main contributions of this article are summarized as follows:
	
	\begin{itemize}
		\item We consider a general scenario of a DT mobile edge network with multiple vehicles and a single VEC server. We analyze two types of delays under this network that are caused by the resource allocation policy.
		\item In this paper, we focus on the problem of resource allocation when twin maintenance is performed simultaneously with on-board tasks, which has been neglected in the existing literature. Therefore, we characterize the two types of delays using the satisfaction function and construct the associated optimization problem taking into account various constraints.
		\item We propose a MADRL-based TMTPRCS algorithm to solve the optimization problem. The effectiveness and superiority of our proposed method is verified by comparing it with other algorithms.
	\end{itemize}
	The remaining part of this article is organized as follows. Section \uppercase\expandafter{\romannumeral2} reviews the related work. Section \uppercase\expandafter{\romannumeral3} introduces the system model and analyzes two types of delays caused by twin maintenance and computing task processing, and then uses satisfaction functions to formulate optimization problems. In section \uppercase\expandafter{\romannumeral4}, we reconstruct the problem using multi-agent MDP and propose a corresponding algorithm. Section \uppercase\expandafter{\romannumeral5} presents simulation results to demonstrate the effectiveness of our proposed algorithm. Section \uppercase\expandafter{\romannumeral6} draws conclusions.
	
	\section{Related work}
	In this section, we first review the relevant work on DT, and then investigate the existing work on the combination of VEC and DT.
	
	\subsection{\textit{Digital Twin}}
	In recent years, there has been a significant amount of research on DTs. In [36], Meysam \textit{et al}. first modeled the BS energy-saving problem as a Markov Decision Process (MDP) to address the significant delay caused to users by not entering sleep mode at the correct time during energy-saving. They also encapsulated the dynamics of the studied system using a DT model and used DT to estimate the risk of decision-making of BS entering sleep mode in advance. In [37], Liao \textit{et al}. addressed the lack of effective coordination in the river area of vehicle ramps and established a DT model for drivers and vehicles, synchronizing data with real-world vehicles. The data was processed using the DT model and the results were returned to both vehicles and drivers, thus solving the sustainability problem of ramp merging. In [38], He \textit{et al}. integrated DT and mobile edge computing into the federated learning framework of heterogeneous cellular networks, and assisted local training of user equipment to reduce latency by deploying DT network models at macro base stations. In [39], Liu \textit{et al}. combined DT with edge collaboration and proposed DT assisted mobile users for task offloading and modeled it as MDP. Then, they decomposed it into two sub models and solved them using decision tree algorithm and dual deep qlearning. In [40], Lu \textit{et al}. integrated DTs with edge networks, proposed DT edge networks, and used deep reinforcement learning to solve the placement and transfer problems of DTs caused by network dynamics and topology changes.
	
	\begin{figure}[!t]
		\centering
		\includegraphics[width=3.2in]{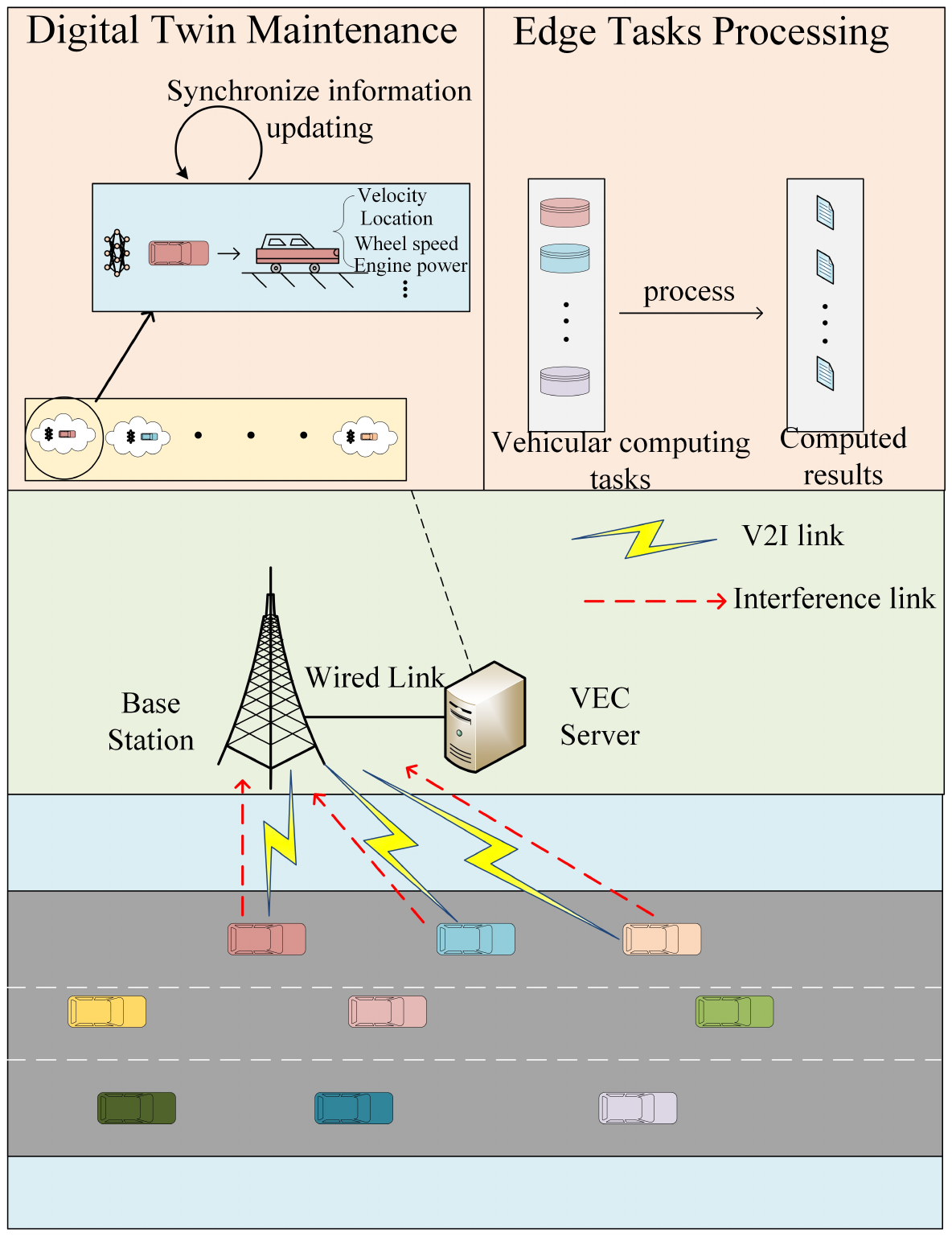}
		\caption{Digital twin mobile edge network.}
		\label{figure1}
	\end{figure}
	
	\subsection{\textit{Digital Twin Vehicular Edge Computing}}
	Recently, there have been some studies combining DTs with VEC. In [26], Dai \textit{et al}. proposed an adaptive DT VEC network that utilizes DT caching content and designs an offloading scheme based on the DRL framework to minimize offloading latency. In [41], Zhang \textit{et al}. combined DTs with artificial intelligence into the VEC network, guiding vehicles to aggregate edge services by deploying digital twins in road side units (RSUs) to minimize offloading costs. In [42], Liao \textit{et al}. considered that in mixed traffic scenarios, the planned operation of autonomous vehicles (CAVs) may be affected by human driven vehicles (HDVs), and developed a driver DT model that was deployed on the server to help CAVs predict the possible lane changing behavior of surrounding HDVs. In [43], Zhang \textit{et al}. proposed a socially aware vehicle edge caching mechanism, which constructs a vehicle social relationship model by deploying a digital twin model of edge caching systems on RSUs and designs an optimal caching scheme based on DRL. In [44], Zheng \textit{et al}. proposed that vehicles in heterogeneous vehicle networks need to improve communication efficiency by selecting different networks, while also facing situations of uneven vehicle distribution and dynamic networks. They established a prediction model in DT to predict the waiting time for vehicles to connect to the network and return the results to the vehicles for decision-making.In [45], Zhao \textit{et al}. established a vehicle DT model in RSUs to learn the global information of the vehicle, thereby assisting clustering algorithms in reducing the scope of vehicle task offloading and achieving the goal of offloading prediction. In [46], Sun \textit{et al}. proposed the Digital Twin Edge Network (DITEN) and established DT models for edge servers and the entire MEC system to provide auxiliary services for mobile user business offloading, thereby minimizing offloading latency. In [47], Li \textit{et al}. integrated unmanned aerial vehicles (UAVs) into VEC and established DT models for RSUs, UAVs, and vehicles in the central controller to manage UAV and RSU resources and assist in offloading vehicular tasks. Although establishing DT models of vehicles in the server of the VEC network can help vehicles perform better task offloading, edge business aggregation, or execute specific traffic behaviors, at the same time, the maintenance of each vehicle's DT model will also consume server resources, which also needs to be considered.
	
	As mentioned above, most of the literature related to digital twins does not take into account the fact that the maintenance of the twin model consumes server computing resources.

	\section{System model}
	As shown in Figure 1, we consider a DT scenario of Mobile Edge Network with $N$ vehicles and a server equipped base station. For clarity, let $\mathbf{N} \stackrel{\Delta}{=} \{1,2,3,..., N\}$ represent the number of vehicles driving on the lane, and its initial position follows the Poisson clustering process. Specifically, these vehicles will travel in one direction along the lane and generate computing tasks during their journey. Due to their limited computing resources, the vehicles will connect to the base station (BS) through the V2I link using a cellular interface to offload the computing tasks to the server for processing. Meanwhile, in order to maintain the digital twin models of these vehicles, the server needs to synchronize information with the vehicles to ensure that the models can accurately reflect the driving conditions of the vehicles. It is noted that we assume that vehicles are always traveling within the coverage area of the current base station. Therefore the migration of the vehicle DT model is not considered.
	
	\subsection{Digital Twin Model}
	The edge server maintains the DT model of vehicle $i$ and the DT model is denoted as
	\begin{equation}
		DT_{i} = \{conf_{i}, His_{i}, \Gamma _i (t)\},
	\end{equation}
	where $conf_{i}$ is configuration information for vehicle $i$, $His_{i}$ is historical information of vehicle $i$, $\Gamma _i (t)$ is the information about the operation of vehicle $i$ in time slot $t$, which can be represented as
	\begin{equation}
		 \Gamma _i (t)=\{{D_i ^{\text{dt}} (t), C_i ^ {\text{dt}} (t), T_i ^ {\text{dt}}}\},
	\end{equation}
	where $D_i ^{\text{dt}} (t)$ is the size of the information, $C_i ^ {\text{dt}} (t)$ represents the CPU frequency required to update the unit size of vehicle $i$ information, and $T_i ^ {\text{dt}}$ is the maximum delay limit. Here, the configuration information $conf_{i}$ mainly refers to the parameters of the vehicle's own hardware, which are fixed. Historical information $His_{i}$, on the other hand, mainly consists of data such as the vehicle's historical mileage as well as historical fuel consumption. Therefore,  these two symbols are not related to $\Gamma _i (t)$.
	
	Note that the DT model developed in this paper is simple. In practical situations, the DT model will not only reflect physical entities, but will likewise reflect network topology relationships. In addition, it has functions such as virtual prediction as well as feedback control.
	
	\subsection{Vehicle Movement Model and Channel Model}
	\subsubsection{Vehicle Movement}
	Similar to [48], a spatial orthogonal coordinate system is established with the base station as the origin in figure 2. This linear movement model is a simulation of the ideal driving condition of a vehicle on a highway with fewer cars. The real situation is much more complex. The positive direction of the x-axis is the direction of the vehicle traveling along the lane, that is, east, the positive direction of the y-axis is south, and the positive direction of the z-axis is along the direction of the base station antenna. Therefore, the coordinate of vehicle $i  \in \mathbf{N}$ located on lane j at time slot t can be represented as $PO_{i} (t)$
	\begin{equation}
		PO_{i}(t)=(X_{i,j}(t),Y_{i,j}(t),Z_{i,j}(t)),i \in \{1,2,3..., N\},
	\end{equation}
	where $X_{i,j}(t)$ is the horizontal ordinate of vehicle $i$ driving on lane $j$ in time slot $t$, $Y_{i,j}(t)$ is the ordinate of vehicle $i$, $Z_{i,j}(t)$ is the vertical coordinate of the vehicle, which we consider as 0.
	
	When the value of each time slot $\tau$ is small enough, the position of the vehicle in each time slot can be approximately considered constant[49]. At the same time, the position of the vehicle in the current time slot $t$ is related to the position in the previous time slot $t-1$. Therefore, horizontal ordinate $X_ {i,j} (t)$ of vehicle $i$ driving on lane $j$ in time slot $t$ can be further represented as
	\begin{equation}
		X_{i,j}(t)=X_{i,j}(t-1) + \tau v_{i},
	\end{equation}
	where $\tau$ is the duration of each time slot and $v_{i}$ is the driving speed of vehicle $i$.
	
	Due to vehicle $i$ drives on lane $j$, we let $w_{0}$ represents the width of each lane, and $L_{0}$ represents the distance between the first lane and BS.Therefore, the ordinate $Y_{i,j}(t)$ of vehicle $i$ can be written as
	\begin{equation}
		Y_{i,j}(t)=L_{0} + j w_{0},j \in (0,1,2,3...J-1),
	\end{equation}
	
	\begin{figure}[!t]
		\centering
		\includegraphics[width=3.2in]{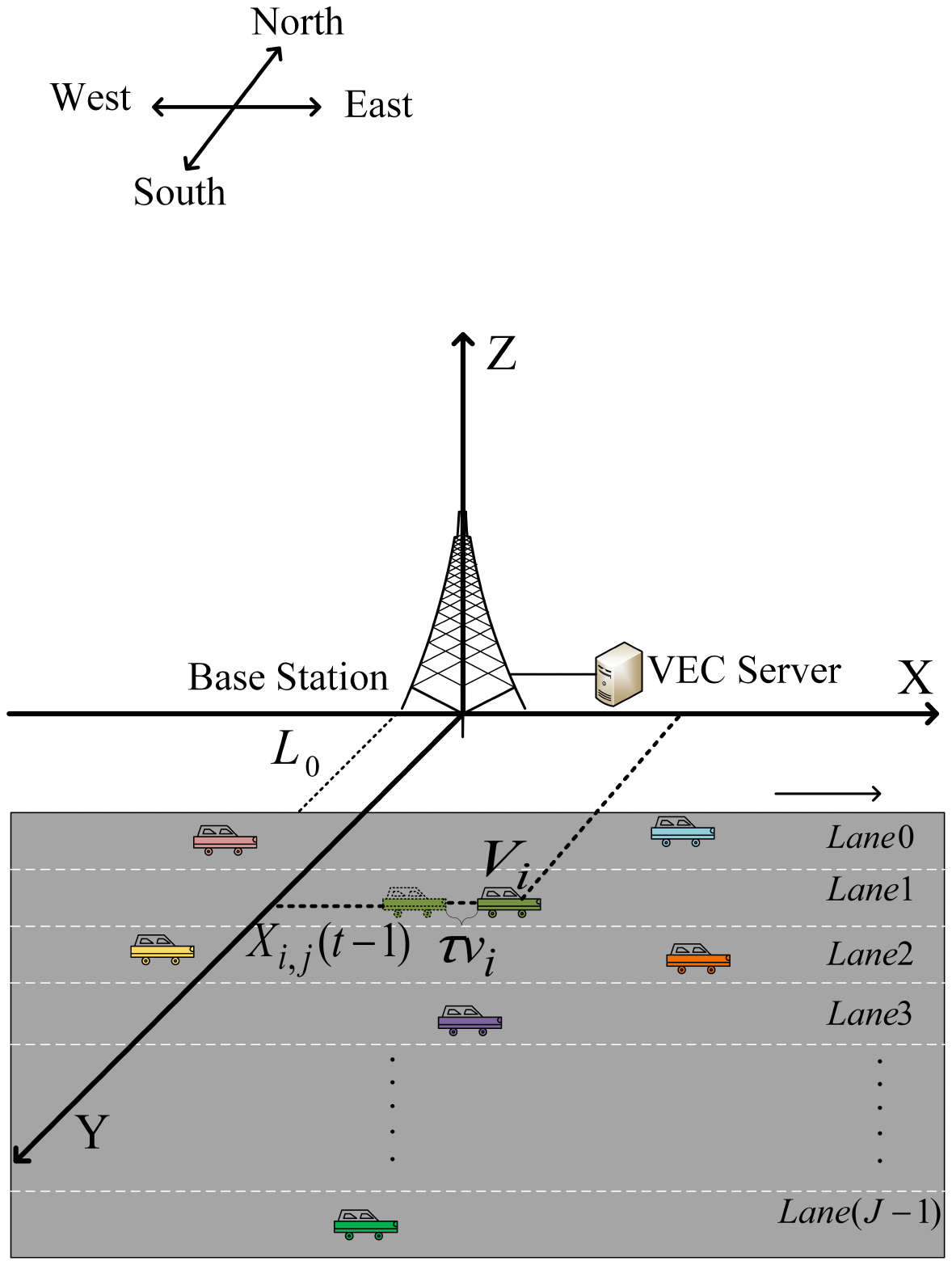}
		\caption{Vehicle movement model.}
		\label{figure12}
	\end{figure}
	
	\subsubsection{Channel Model}
	We use Orthogonal Frequency Division Multiple Access (OFDMA) in DTMEN for communication between vehicles and base stations, so we don't consider interference between vehicles[50][51]. At the current time slot $t$, the channel gain between vehicle $i$ and BS, $g_{i, B} ^ t$ is modeled as[52]
	\begin{equation}
		g_{i, B} ^ t = |h_{i,B} ^t|^2 v_{i,B} ^t, t=0,1,2,...,
	\end{equation}
	where $v_{i,B} ^t$ is the large-scale fading component between vehicle $i$ and BS, which is composed of path loss and shadow distribution, where path loss is calculated as $128.1 + 37.6\log_{10}(dis(v_i, BS))$ and shadowing obeys the log-normal distribution. $dis(v_i,\text{BS})$ is the distance between vehicle $i$ and BS. $h_{i,B} ^t$ is the small-scale path fading component between vehicle $i$ and BS at time slot $t$, which follows a circularly symmetric complex Gaussian distribution with unit variance, i.e $h_{i,B}\sim\mathcal{CN}(0,1)$. We use a first-order Gaussian Markov process to describe small-scale path fading. Therefore, the update of $h_{i,B} ^t$ can be represented as
	\begin{equation}
		h_{i,B} ^t = \kappa h_{i,B} ^{t-1} + e_{i,B}^t,
	\end{equation}
	where $e_{i,B}^t$ is the channel innovation process with distribution $\mathcal{CN}(0,1-\kappa^2)$. the correlation coefficient $\kappa$ is denoted as $\kappa = J_0(2\pi f_d \tau)$, where $J_0(\bullet)$ is the zeroth order Bessel function. $f_d$ is the maximum Doppler frequency, which is calculated as $f_d=\frac{v_i f_c}{c}$, where $f_c$ is the carrier frequency, $c$ is the transmission rate of electromagnetic waves.
	
	\subsection{Communication Model and Computing Model}
	\subsubsection{Communication Model}
	As shown in Figure 1, the communication between the vehicle and the base station includes information updates for vehicle twin maintenance, uplink communication for task transmission with the same server, and downlink communication for returning calculation results to the vehicle. It should be noted that since the size of the results processed by the digital twin edge server is much smaller than before, and the communication rate of the downlink is much larger. Therefore, we ignore the delay caused by downlink communication. According to Shannon's theorem, we can calculate the uplink communication rate between the vehicle and the base station as follows[53]
	\begin{equation}
		r_{i,B} ^t = W \log(1+\frac{p_i g_{i, B} ^ t}{\sigma ^2}),
	\end{equation}
	where $W$ is the bandwidth for wireless transmission between vehicle $i$ and BS, $p_i$ is the transmit power of vehicle $i$, $\sigma ^2$ is the noise between vehicle $i$ and BS.
	
	\subsubsection{Computing Model}
	After the DT model is created in the VEC server, DT model maintenance is important. Digital twin maintenance refers to the process of synchronizing information between the twin model and its physical entity in real time, i.e., the digital twin model self-updates itself by profitably processing real-time information transmitted by its physical entity. In this paper, we mainly focus on the process of processing information by the twin model. The updated information needs to be transmitted from vehicle $i$ to the server equipped with the base station first, and then send a request signal to its DT model for requesting a portion of the server resources to maintain the model. Here, according to the colocation restrictions of [54], the maintenance of digital twins on vehicle $i$ is also related to the total number of digital twins it needs to maintain on its server. In addition, we record the total amount of computing resources on the server as $F$, and the computing resources requested by vehicle $i$ for digital twin maintenance as $f_i ^ {\text{dt}}$. Therefore, the time $t_i ^ {\text{dt}}$ required to complete the digital twin maintenance of vehicle $i$ can be calculated as
	
	\begin{equation}
		t_i ^ {\text{dt}} = \frac{D_i ^{\text{dt}} (t) }{r_{i,B} ^t} + \frac{	N D_i ^{\text{dt}} (t) C_i ^ {\text{dt}} (t)}{f_i ^ {\text{dt}}} ,
	\end{equation}
	Digital twin maintenance is modeled here as the process by which the data used by vehicles to maintain their models is synchronized to the digital models in the server. At the same time the process is influenced by the total number of twin models that the server has to maintain. This process is embodied in the algorithm designed in Section IV in that it acts on the rewards at each time step of the intelligence, which in turn directly affects policy learning.
	
	Vehicles may generate computing tasks while driving, and due to their insufficient computing power, they may choose to offload the computing tasks to edge servers for processing. We represent the computing task generated by vehicle $i$ at time slot $t$ as $ \zeta _i (t)=\{{D_i ^ {\text{tk}} (t), C_i ^ {\text{tk}} (t), T_i ^ {\text{tk}}}\}$, where $D_i ^ {\text{tk}} (t)$ is the computing task size generated by vehicle $i$ at time slot $t$, $C_i ^ {\text{tk}} (t)$ is the CPU frequency required to compute unit size tasks, and $T_i ^ {\text{tk}}$ is the maximum delay limit for processing computing tasks. Similar to the previous maintenance of digital twins, the vehicle $i$ will first offload the computing task to the server, and then request computing resources to handle the computing task. Therefore, the time $t_{i} ^ {\text{tk}}$ required to complete the calculation task, which can be calculated as
	\begin{equation}
		t_{i} ^{\text{tk}} = \frac{D_i ^ {\text{tk}} (t) }{r_{i,B} ^t} + \frac{D_i ^ {\text{tk}} (t) C_i ^ {\text{tk}} (t)}{f_i ^ {\text{tk}}},
	\end{equation}
	where $f_i ^ {\text{tk}}$ represents the computing resources requested by vehicle $i$ for processing computing tasks transmitted to the server.
	
	It should be noted that the maintenance and computing tasks of the digital twin of vehicle $i$ are carried out simultaneously, so the sum of the computing resources used for twin maintenance and computing tasks processing for all vehicles can't exceed the total computing resources $F$ of the server, that is, $\Sigma_1 ^ N  (f_i ^ {\text{dt}}+f_i ^ {\text{tk}}) \le F$.
	
	\subsection{Optimization Problem }
	Due to the fact that vehicle $i$ simultaneously performs digital twin maintenance and offloading of computing tasks at time slot $t$, vehicle $i$ will request computing resources for both, namely $f_i ^ {\text{dt}}$ and $f_i ^ {\text{tk}}$, respectively. By using different resource request strategies, different information synchronization delays and task computing delays can be obtained, which in turn affects the maintenance delay $t_{i, B} ^ {\text{dt}}$ of digital twins and the processing delay $t_{i, B} ^ {\text{tk}}$ of computing tasks. According to [44], we define a satisfaction function $Q$ to demonstrate the satisfaction level of digital twin maintenance delay and computing task processing delay for vehicle $i$ under a certain resource request strategy $\omega _i$, respectively. Therefore, the satisfaction function $Q_i ^ {\text{dt}} (\omega _i)$ for the maintenance delay of vehicle $i$ digital twin and the satisfaction function $Q_i ^{\text{tk}} (\omega _i)$ for the processing delay of the computational task for vehicle $i$ can be respectively represented as
	\begin{equation}
		Q_i ^ {\text{dt}} (\omega _i) = 1-\ln(1+\frac{t_i ^{\text{dt}}}{T_i ^{\text{dt}}}),
	\end{equation}
	\begin{equation}
		Q_i ^ {\text{tk}} (\omega _i) = 1-\ln(1+\frac{t_i ^{\text{tk}}}{T_i ^{\text{tk}}}),
	\end{equation}
	
	The satisfaction function reflects the level of satisfaction with the current resource allocation strategy, as the twin maintenance latency and task processing latency are higher, the value of the satisfaction function is lower, which indicates that the current strategy is poor. Therefore, through (9) and (10), we can evaluate the resource utility $U_i (\omega _i)$ obtained by vehicle $i$ under a resource allocation strategy $\omega _i$
	\begin{equation}
		U_i (\omega _i) = \rho Q_i ^ {\text{dt}} (\omega _i) + (1-\rho)Q_i ^ {\text{tk}} (\omega _i),
	\end{equation}
	where $\rho$ is the weighting factor with $0<\rho<1$ between twin maintenance and computational task processing, used to measure the importance of the two tasks. The weight parameters are set to be fixed here because the vehicle environment considered in this paper is a smooth state in an ideal environment. However, it is important to note that dynamic weights are more suitable to cope with emergency mission situations.
	
	Our goal is to maximize the resource utility $U$ obtained by each vehicle while maintaining a fixed total server resource and meeting the maximum latency conditions for twin maintenance and computing tasks. Therefore, the optimization problem under consideration can be formulated as
	\begin{subequations}\label{P1}
	\begin{equation}
		P1:{\rm{}}\mathop {\max}_{i \in N}\quad U_i (\omega _i)\\
	\end{equation}
	\begin{equation}
		{\rm{			}}s.t.\quad{\rm{	}}t_i ^{\text{dt}} \le T_i ^{\text{dt}},\\
	\end{equation}
	\begin{equation}
	\qquad{\rm{		}}t_i ^{\text{tk}} \le T_i ^{\text{tk}},\\
	\end{equation}
	\begin{equation}
		\qquad\qquad\quad\sum_{i=1} ^N (f_i^{\text{dt}} + f_i ^{\text{tk}})\le F,\\
	\end{equation}
	\begin{equation}
		\qquad\qquad\quad p_i \le p_i ^{max},\forall i\in N,\\
	\end{equation}
	\end{subequations}
	where $p_i ^ {max}$ is the maximum transmit power of vehicle i. Constraint (14b) requires that the maintenance time of digital twins cannot exceed the maximum delay tolerance. Constraint (14c) requires that the processing time of computing tasks cannot exceed the maximum delay tolerance. Constraint (14d) requires that the sum of computing resources consumed by all vehicles cannot exceed the server's own computing resource capacity. Constraint (14e) means that the transmission power of each vehicle cannot exceed the maximum value.
	
	\section{Proposed solution}
	Each vehicle in the scene will simultaneously perform digital twin maintenance and offloading of computing tasks to the server. The computing resources requested by a vehicle to complete these two tasks will also affect the resources obtained by other vehicles, thereby affecting the final utility. We transformed the problem into a multi-agent MDP and proposed a TMTPRCS algorithm to solve it, while analyzing the complexity of the algorithm. The algorithm refers to MADDPG and is partially optimized based on it. Specifically, the traditional MADDPG algorithm relies on injecting Gaussian noise for exploration, which can be inefficient and lacks dynamic adaptation. In contrast, TMTPRCS employs an adaptive $\epsilon$-greedy algorithm that dynamically adjusts throughout the training process, promoting exploration during the early stages and gradually shifting towards exploitation as training progresses. Unlike conventional greedy algorithms, TMTPRCS ensures that effective actions and corresponding knowledge are not lost during the exploration phase.
	
	\subsection{Multi-agent MDP transformation of problems}
	The multi-agent MDP problem can be described by using a five tuple $(\mathcal{N, S, A, P, R})$, where $\mathcal{N, S, A, P, R}$ in the tuple represent the set of agents, state space, action space, state transition probability, and reward function, respectively.
	\subsubsection{\textbf{Agent Set } \bm{$\mathcal{N}$}}
	In order to achieve the goal of maximizing vehicle utility, each vehicle will act as an agent to learn resource allocation for digital twin maintenance and computing tasks. So, we let the agent set $\mathcal{N}=\{1,2,3,... N\}$
	\subsubsection{\textbf{State Space} \bm{$\mathcal{S}$}}
	The state $s_n (t)$ of agent $n$ at moment $t$ can be described by updating information, computing tasks, vehicle position, speed, and channel gain, which can be represented as $s_n (t) = \{\Gamma _n (t), \zeta_n (t), PO_n (t), v_n, g_{n,B}^t\}$. The state of the entire system $s(t)$ can be expressed as $s (t)=\{s_n (t)\} _ N$.
	\subsubsection{\textbf{Action Space} \bm{$\mathcal{A}$}}
	The behavior $a_ n (t)$ taken by the agent $n$ includes request for resources for twin maintenance $f_ n ^ {\text{dt}}$ and computing task processing $f_ n ^ {\text{tk}}$. Therefore, the behavior $a_ n (t)$ can be expressed as $a_n (t)=\{f_n ^{\text{dt}}, f_n ^{\text{tk}}\}$. The action space $a(t)$ for all agents can be represented as $a (t)=\{a_ n (t)\}_N$.
	\subsubsection{\textbf{State Transition Probability} \bm{$\mathcal{P}$}}
	The state transition probability $p_n (t)$ describes the probability that at each decision epoch $t$, the agent $n$ executes action $a_n (t)$ in the current state $s_n (t)$ and then transfers to the next state $s_n (t+1)$, that is, $p_n (s_n (t+1); s_n (t), a_n (t))$.
	\subsubsection{\textbf{Reward Function} \bm{$\mathcal{R}$}}
	The reward function will return the corresponding reward value or a penalty value when the agent $n$ takes action in a given state. By considering the optimization problem itself and its constraints, we let the Utility $U$ serve as the reward function, that is, the reward function $r_n (t)$ of agent $n$ is
	\begin{equation}
		r_n (t) =  \rho Q_n ^ {\text{dt}} (\omega _n) + (1-\rho)Q_n ^ {\text{tk}} (\omega _n),
	\end{equation}
	on this basis, the long-term discount reward we can receive is
	\begin{equation}
		R_n (t) = \sum_{t_0} ^t {\gamma_n r_n (t)},
	\end{equation}
	where $t_0$ is the previous epoch, $\gamma_ n$ is the discount factor, with a range of values $\gamma_ n \in [0,1]$, which represents the degree to which past rewards have an impact on the rewards at the current epoch $t$.
	
	By maximizing the long-term cumulative reward of each agent, i.e., $\max R_n(t)=\sum_{t_0}^t {\gamma_n r_n(t)}$, we can then obtain the optimal resource allocation strategy.
	
	\begin{figure*}[htbp]
		\centering
		\includegraphics[width=5.0in]{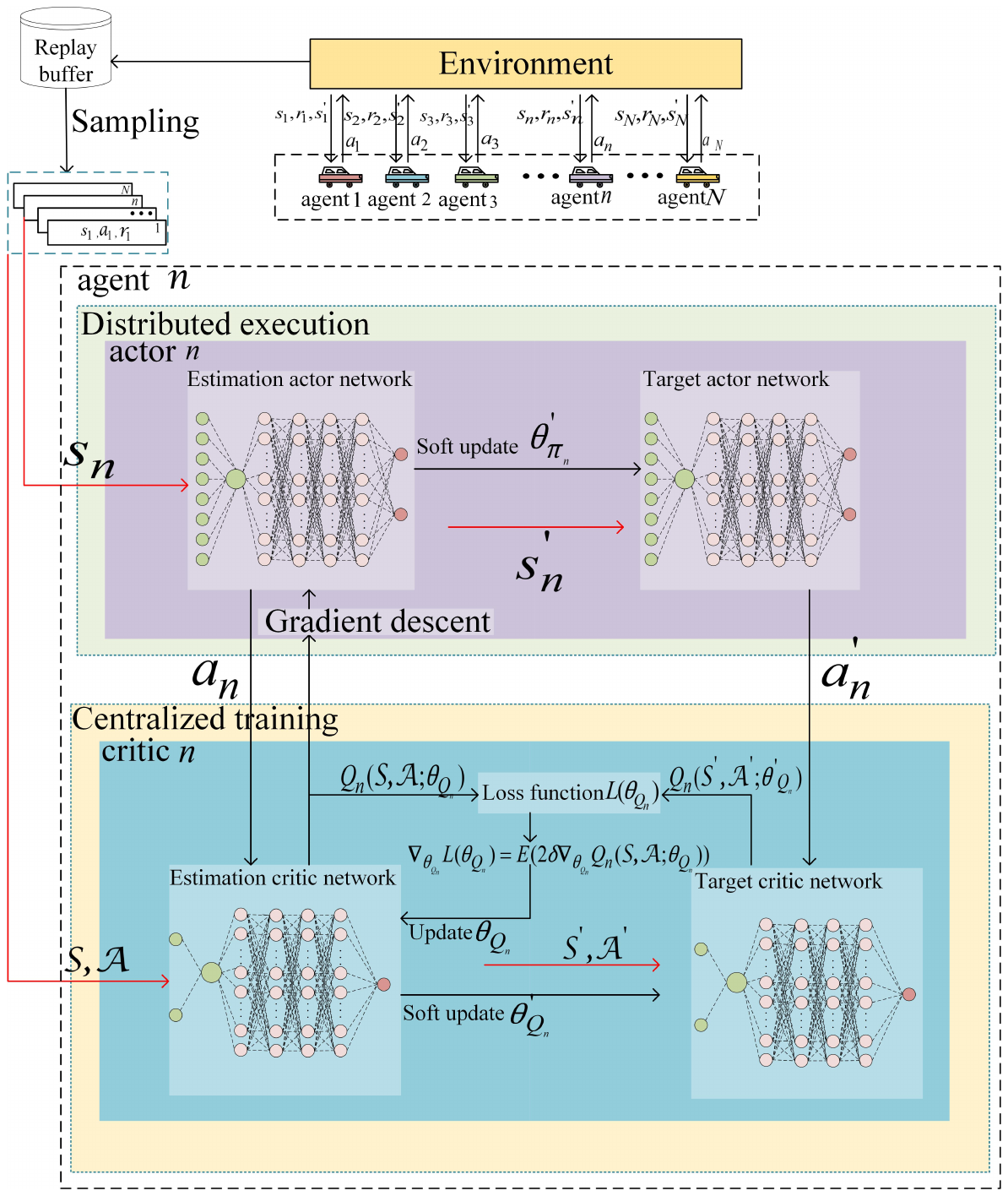}
		\caption{Structure of TMTPRCS}
		\label{figure3}
	\end{figure*}
	
	\subsection{Algorithm Design}
	As shown in Figure 3, for the structure of TMTPRCS algorithm, actor-critic framework is employed, where the actor is used to generate the actions that the agent needs to perform, while the critic is responsible for guiding the actor network to generate better actions. Here, the actor consists of two parts, where the estimation actor network is responsible for training, and the target actor network is responsible for executing the action. Similarly, critic also includes two parts, namely the estimation critic network and the target critic network, both of which are used to evaluate the actions of agents. For this reason, the actor network adopts a policy based deep neural network, while the critic uses a value based deep neural network. In addition, considering the dynamic changes in the environment, we adopt a centralized training and distributed execution strategy, where the critic is centrally trained by the server and the actor is executed by each vehicle in a distributed manner. It is noted that centralized training is done in an offline manner and distributed execution is online.
	
	\subsubsection{actor}
	We construct the actor network into three layers: input layer, fully connected layer, and output layer. The fully connected layer is composed of three hidden layers and one softmax layer. The first two hidden layers use the rectified linear unit (ReLu) function as the activation function, and the third hidden layer uses the tangent (tanh) function as the activation function. By constructing an actor network in this way, the input state can be transformed into all possible actions, namely resource allocation strategies. Since the actor network is divided into two parts, we will explain the estimation actor network and the target actor network separately.
	
	For the estimation actor network of agent $n$, its input is the current state $s_n (t)$, including the synchronization information required for twin maintenance $\Gamma_ n (t)$, computational tasks generated by vehicles $\theta_ n (t)$, vehicle position $PO_n (t)$, vehicle speed $v_n$, and channel gain $g_{n,B}^t$. The current state is input into the fully connected layer, processed by three hidden layers, and output as the probability of all possible actions. After passing through the softmax layer, the total probability of all actions is set to 1. Finally, agent $n$ selects one of them as the final resource allocation strategy to be executed.
	
	Similar to estimation actor network, the input of target actor network is the next state $s_n (t+1)$, and the corresponding output is the next action $a_n (t+1)$ processed by the fully connected layer. It is should be noted that although the structure of target actor network and estimation actor network are with the similar structure, their parameters $\theta^{'} (\pi_n)$ and $\theta (\pi_n)$ are different respectively, where the former is the target actor network parameter and the latter is the estimation actor network parameter.
	
	\subsubsection{critic}
	Similar to the actor network, the structure of the critic network includes an input layer, a fully connected layer, and an output layer. The difference is that only the first three layers in the fully connected layer have activation functions, namely Relu layer, Relu layer, and Tanh layer.
	
	For the estimation critic network of agent $n$, its input is the state and actions of all agents at the current epoch, namely $\mathcal{S}$ and $\mathcal{A}$. After the current input is processed by the fully connected layer, the Q value is obtained. For each decision epoch $t$, the Q value of agent $n$ can be defined as $Q_n(\mathcal{S},\mathcal{A};\theta_{Q_n})=\mathbb{E}_{\pi} [R_n (t);s(t),a(t);\theta_{Q_n}]$.
	
	For the target critic network of agent $n$, similar to the estimation critic network, the input is the state and action of all agents at the next epoch, and after being processed by the fully connected layer, the corresponding $Q$ value \( Q_n ^ {'}(\mathcal{S} ^{'}, \mathcal{A} ^{'};\theta_{Q_n} ^{'}) \) is obtained. It should be noted that the structure of the target critic network is the similar to that of the estimation critic network, but their parameters are different, i.e. $\theta^{'}_ {Q_n}$ and $\theta_{Q_n}$.
	
	\begin{algorithm}[!t]
		\caption{Training Stage of TMTPRCS Algorithm.}\label{alg:alg1}
		\renewcommand{\algorithmicrequire}{\textbf{Input:}}
		\renewcommand{\algorithmicensure}{\textbf{Output:}}
		\begin{algorithmic}[1]
			\REQUIRE{$N, D_n ^{\text{dt}}, C_n ^{\text{dt}}, T_n ^{\text{dt}}, D_n ^{\text{tk}}, C_n ^{\text{tk}}, T_n ^{\text{tk}}, v_n, g_{n,B} ^t$ for $n=1,...,N$;}
			\ENSURE{$\theta_{\pi_n} ^{'}$ for $n=1,...,N$}
			\STATE Initialize discount factor $\gamma$ and parameter update rate $\eta$; 
			\STATE Random initialize the $\theta_{Q_n}, \theta_{Q_n} ^{'}, \theta_{\pi_n}, \theta_{\pi_n}^{'}$ for $n=1,...,N$;
			\FOR{$k$ eposide from 1 to $K$}
			\FOR{$n$ agent from 1 to $N$}
			\STATE Initialize $s_n (t)$;
			\STATE Input $s_n (t)$ to estimation actor network, and get $a_n (t)= \pi_n (s_n(t);\theta_{\pi_n})$;
			\STATE Execute $a_n(t)$ based on $s_n(t)$, obtain $r_n(t)$ and transfer to $s_n(t+1)$;
			\STATE Store $(s_n(t), a_n(t), r_n(t), s_n(t+1))$ as an experience in the Replay Buffer;
			\STATE Input $\mathcal{S}$ and $\mathcal{A}$ to estimation critic network and compute $Q_n(\mathcal{S},\mathcal{A};\theta_{Q_n})$;
			\STATE Input $\mathcal{S^{'}}$ ,$\mathcal{A^{'}}$ to target critic network and compute $Q_n ^{'}(\mathcal{S^{'}},\mathcal{A^{'}};\theta_{Q_n}^{'})$;
			\STATE Calculate Q value, temporal difference $\delta$ and loss function $L(\theta_{Q_n})$;
			\STATE Update $\theta_{Q_n}$ by stochastic gradient descent;
			\STATE Input $s_n(t)$ to estimation actor network and obtain $a_n(t)=\pi_n (s_n(t);\theta_{\pi_n})$;
			\STATE Input $s_n(t+1)$ to target actor network and obtain $a_n(t+1)=\pi_n^{'}(s_n(t+1);\theta_{\pi_n}^{'})$;
			\STATE Update $\theta_{\pi_n}$ by gradient descent;
			\STATE Update $\theta_{Q_n}^{'}$ and $\theta_{\pi_n}^{'}$ , respectively;
			\ENDFOR
			\ENDFOR
		\end{algorithmic}
		\label{alg1}
	\end{algorithm}
	
	\subsection{Algorithm Training}
	The centralized training of the TMTPRCS algorithm is implemented on the server, as summarized in Algorithm 1. Specifically, since the server manages the critic network of each agent, the status and actions of all agents can be obtained, meaning that the information of all agents is observable. In the centralized training phase, the server first obtains the states and actions of all agents. By utilizing this information, the server can train the estimation critical network for each agent, thereby achieving the goal of maximizing the Q value. For a single agent $n$, its Q-value $Q_n (\mathcal{S}, \mathcal{A}; \theta_ {Q_n})$ can be updated through the Bellman equation[55], which is given by $Q_n (\mathcal{S},\mathcal{A};\theta_{Q_n})=R_n (t)+\gamma_n \max Q_n^{'}(\mathcal{S}^{'},\mathcal{A}^{'};\theta^{'} _{Q_n})$.
	The time difference error can be calculated as $\delta = Q_n^{'}(\mathcal{S}^{'},\mathcal{A}^{'};\theta^{'} _{Q_n}) - Q_n(\mathcal{S},\mathcal{A};\theta_{Q_n})$. Therefore, the loss function can be expressed as $L(\theta_{Q_n})=\mathbb{E}(\delta ^2)$. To minimize the loss function $L(\theta_{Q_n})$, we update the parameters $\theta_{Q_n}$ by using a stochastic gradient descent algorithm, which is represented as $\nabla_{\theta_{Q_n}} L(\theta_{Q_n}) = \mathbb{E}(2\delta \nabla_{\theta_{Q_n}} Q_n(\mathcal{S},\mathcal{A};\theta_{Q_n}))$.
	
	Unlike critic networks, due to the distributed execution of actor networks and the need for agents to take actions based on local observations, actor networks are deployed on each vehicle. The parameters of the actor network are updated through gradient descent
	\begin{equation}
		\nabla_{\theta_{\pi_n}} L(\theta_{\pi_n}) \approx \mathbb{E}[\nabla_{\theta_{\pi_n}} \log \pi_n (s_n (t );\theta_{\pi_n}) Q_n (\mathcal{S},\mathcal{A};\theta_{Q_n})],
	\end{equation}
	where $\pi_n (s_n (t); \theta_ {\pi_n})$ represents the strategy for executing actions in the current state.
	
	To ensure the stability of the entire training process, we adopt a soft update approach to update the parameters of the target actor network and target critic network. The specific update methods are $\theta_{Q_n} ^{'} = \eta \theta_{Q_n} + (1-\eta)\theta_{Q_n} ^{'}$ and $\theta_{\pi_n} ^{'} = \eta \theta_{\pi_n} + (1-\eta)\theta_{\pi_n}^{'}$, respectively, where $\eta$ is parameter update rate, while $\eta \in [0,1]$.
	
	\begin{algorithm}[!t]
		\caption{Executing Stage of TMTPRCS Algorithm.}\label{alg:alg2}
		\renewcommand{\algorithmicrequire}{\textbf{Input:}}
		\renewcommand{\algorithmicensure}{\textbf{Output:}}
		\begin{algorithmic}[1]
			\REQUIRE{$N, \theta_{\pi_1}^{'}$, $\theta_{\pi_2}^{'}$,..., $\theta_{\pi_N}^{'}$;}
			\ENSURE{$a_n(t)$ for $n=1,...,N$}
			\FOR{agent $n$ from 1 to $N$}
			\STATE Input $\theta_{\pi_n}^{'}$ to the target actor network of agent $n$;
			\STATE Agent $n$ observe the environment to get current state $s_n(t)$;
			\STATE Choose a probability $Pr_n \in [0,1]$;
			\IF{$Pr_n \le \epsilon$}
			\STATE Randomly select an action $a_n(t)\in \mathcal{A}$;
			\ELSE 
			\STATE Compute the value of action by $a_n(t)=\pi_n^{'}(s_n
			(t);\theta_{\pi_n}^{'})$;		
			\STATE Select the action with the highest Q value;
			\ENDIF
			\ENDFOR
			
		\end{algorithmic}
		\label{alg2}
	\end{algorithm}	
	
	\subsection{Algorithm Execution}
	After the centralized training is completed, distributed actions need to be executed on each vehicle based on local observations. Specifically, agent $n$ downloads the results of centralized training and inputs them into the actor network. Then, agent $n$ observes and obtains the local state $s_n (t)$, and executes action $a_n (t)$ based on policy $\pi_n$ to obtain a reward $r_n (t)$. In addition, due to a lack of experience in the early stages of execution, agent $n$ randomly selects actions to explore. When experience is sufficient, it will execute actions to maximize rewards. Therefore, in order to achieve a balance between exploration and exploitation, we adopt the $\epsilon$-greedy algorithm. In addition, we have optimized $\epsilon$ so that it can be used to maintain a balance between exploration and exploitation as training progresses. It can be expressed as $\epsilon=(1-\lambda)^K \epsilon_0$, where $\lambda$ is rate of descent, $\epsilon_0$ is initial value of $\epsilon$. The detailed process is as summarized in Algorithm 2.
	\subsection{Algorithm Complexity Analysis}
	Similar to the analysis process in [30], the computational complexity of algorithms is mainly related to the structure and parameters of neural networks. Due to the fact that both the actor network and the critic network of the TMTPRCS algorithm use DNN, the calculation of complexity needs to be based on the analysis of DNN. We let $Ln$ represent the number of layers in DNN, and let $G_l$ represent the number of neurons in the $l$-th layer. Therefore, we can obtain that the computational complexity of the actor network and the critic network is $\mathcal{O} (X_a)=\mathcal{O} (X_c)=\mathcal{O} (\sum _ {l=1} ^ {Ln} G_l G_ {l+1})$.
	
	Due to our algorithm adopting a centralized training and distributed execution approach, we still need to analyze these two parts separately. For centralized training, each agent has $E$ experiences stored in the experience replayer, and agent $n (n \in {1,2,..., N}) $needs to train $K$ iteration cycles. Therefore, the computational complexity of actors and critics is $\mathcal{O}_a (X_a K E ^ N)$ and $\mathcal{O}_c (X_c K E ^ N)$, respectively. For distributed execution, agent n only needs to observe the state and then execute actions through the actor network, so the complexity of the actor is $\mathcal{O}_a (X_a)$.
	
	\section{Simulation results}
	In this section, to evaluate the performance of the TMTPRCS algorithm on the scenario considered in this paper, we conducted simulation experiments and compared it with other algorithms to analyze its superiority.
	
	\subsection{Experimental Setup}
	The simulation experiment in this article was implemented using Python 3.11.0 and Pytorch. In terms of environmental settings, the experimental parameters are listed in Table $\mathrm{I}$. The number of lanes $J$ is set to 3. The distance $L_0$ between first lane and BS and the width $\omega_0$ of each lane are 8m and 3.75m, respectively. The duration $\tau$ of each time slot is set to 100ms. The bandwidth $W$ between vehicle $i$ and BS is set to 20Mhz, the transmission power of vehicle $i$ is set to 23dBm and the noise power $\sigma^2$ is -110dBm. The carrier frequency $f_c$ is 2Ghz. The speed $v_i$ of vehicle $i$ randomly sample between [10,15]m/s. We assume that the information required for twin maintenance is the vehicle control task, whose size $D_i ^{\text{dt}}$ and time constraints $T_i ^{\text{dt}}$ are set to uniformly sample between [10,200] bytes and 100ms, respectively. The size $D_i ^{\text{tk}}$ of the computing task generated by the vehicle is set to uniformly sample between [1000, 1500]bytes, and its time limit $T_i ^{\text{tk}}$ is 500ms. The cpu frequency $C_i ^ {dt}/C_i ^ {tk}$ required per unit byte is set to 0.25Mhz/byte.  In terms of DNN, for actor networks, the number of neurons in the first two hidden layers of the three fully connected layers is 300 and 100, respectively, and the number of neurons in the third hidden layer is the action dimension. For the critic network, the number of neurons in the three hidden layers is 300, 100 and 1, respectively. The learning rates for the actor network and critic network are $lr_a=10 ^ {-4}$ and $lr_c=10 ^ {-3}$, respectively. The discount factor $\gamma_ n$ and parameter update rate $\eta$ are set to 0.95 and 0.01. It is noted that simulation parameters are fixed to ensure fair baseline comparisons. We acknowledge that the sensitivity of TMTPRCS to variations in environmental and algorithmic parameters remains an open question, and plan to conduct a detailed sensitivity analysis in future work to evaluate robustness under diverse network settings.
	
	\begin{table}[!t]
		\caption{Experimental parameter settings\label{tab:table1}}
		\renewcommand\arraystretch{1.0}
		\centering
		\begin{tabular}{|c|c|}
			\hline
			\textbf{Parameter} & \textbf{Value} \\
			\hline
			$J$ & 3 \\
			\hline
			$L_0$ & 8m \\
			\hline
			$\omega_0$ & 3.75m \\
			\hline
			$\tau$ & 100ms \\
			\hline
			$W$ & 20Mhz \\
			\hline
			$p_i$ & 23dBm \\
			\hline
			$\sigma^2$ & -110dBm\\
			\hline
			$f_c$ & 2Ghz \\
			\hline 
			$v_i$ & [10,15]m/s\\ 
			\hline
			$D_i ^{\text{dt}}$ & [10,200]bytes \\
			\hline
			$D_i^{\text{tk}}$ & [1000, 1500]bytes \\
			\hline
			$T_i^{\text{dt}}$ & 100ms \\
			\hline
			$T_i^{\text{tk}}$ & 500ms \\
			\hline
			$C_i ^ {dt}, C_i ^ {tk}$ & 0.25Mhz/Byte \\ 
			\hline
			$lr_a$ & $10^{-4}$ \\
			\hline
			$lr_c$ & $10^{-3}$ \\
			\hline
			$\gamma_n$ & 0.95 \\
			\hline
			$\eta$  & 0.01 \\
			\hline
		\end{tabular}
	\end{table}
	
	\begin{figure}[h]
		\centering
		\includegraphics[width=\columnwidth]{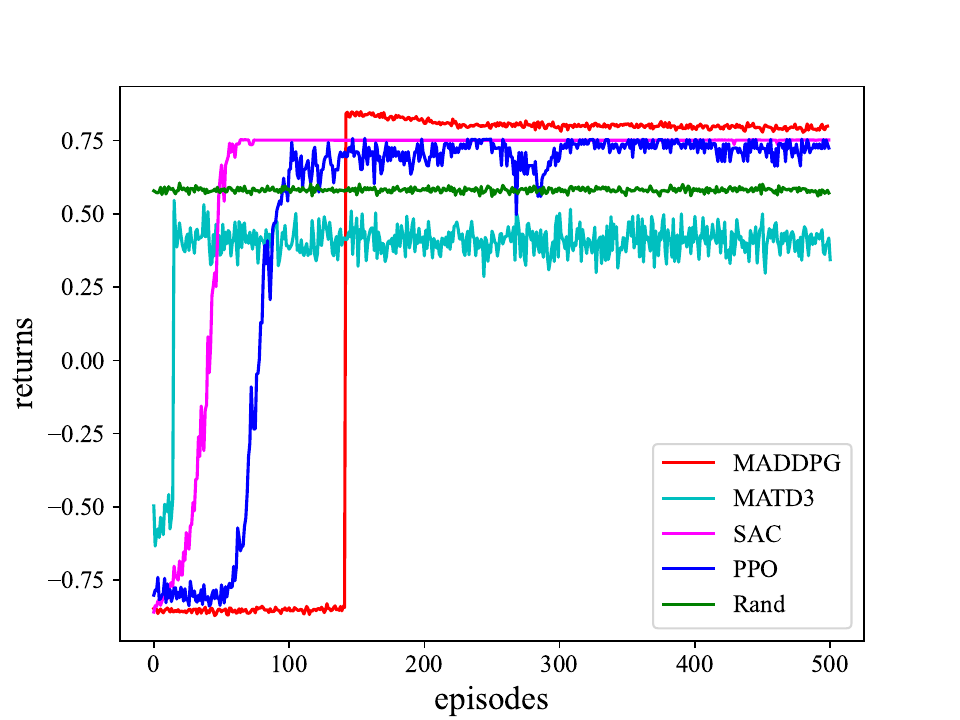}
		\caption{Average rewards under different algorithms:$N=7$}
		\label{figure4}
	\end{figure}
	
	\subsection{Algorithm Comparison}
	We compared four algorithms, namely Multi-Agent Twin Delayed Deep Deterministic(MATD3) algorithm, Policy Gradient Soft Actor-Critic(SAC) algorithm, Proximal Policy Optimization(PPO) algorithm, and Random algorithm.
	\begin{itemize}
		\item MATD3: A Multi-agent Algorithm for improving stability and performance by introducing dual Q networks and delayed updates of target policies.
		\item SAC: An algorithm based on maximum entropy reinforcement learning aims to maximize the cumulative reward while maximizing the entropy of the strategy to encourage exploration.
		\item PPO: An algorithm that ensures training stability by limiting the magnitude of policy updates.
		\item Random: We make a simple modification to the Random algorithm to satisfy the constraint 14d. Specifically, Random algorithm is the process of first dividing the total computing resources by the number of agents at each decision time $t$, and then randomly allocating computing resources for twin maintenance and processing computing tasks to the corresponding agents within each average component.
	\end{itemize}
	
	\begin{figure*}[!t]
		\centering
		\begin{subfigure}[Resource utilization]{
				\centering
				\includegraphics[width=3.0in]{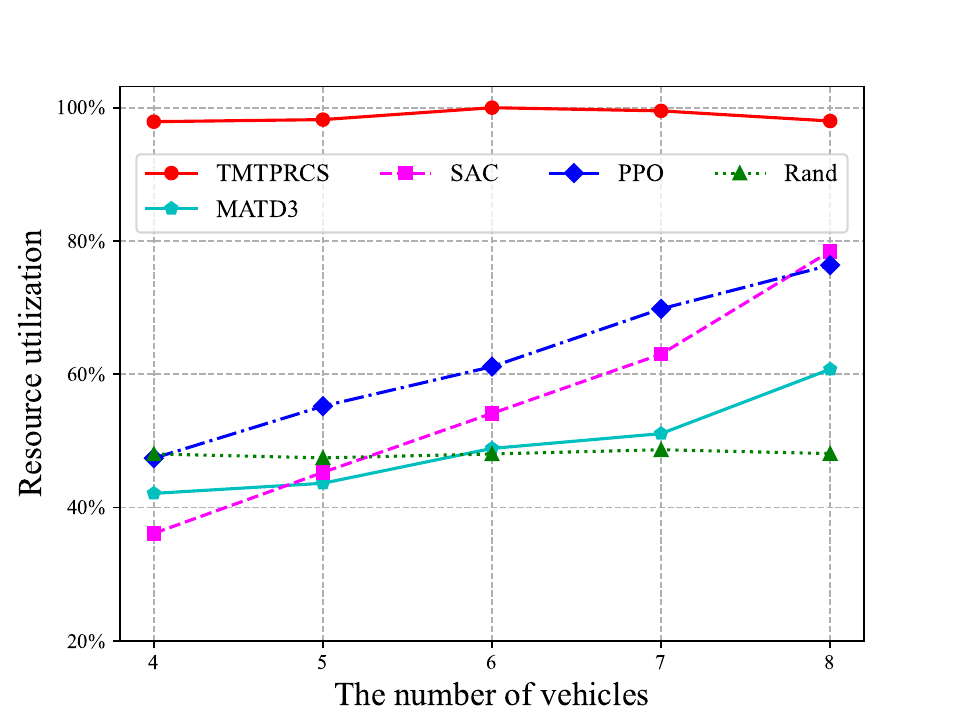}
				\label{figure5.a}}
		\end{subfigure}
		\centering
		\begin{subfigure}[Schedulable resources for twin maintenance]
			{
				\centering
				\includegraphics[width=3.0in]{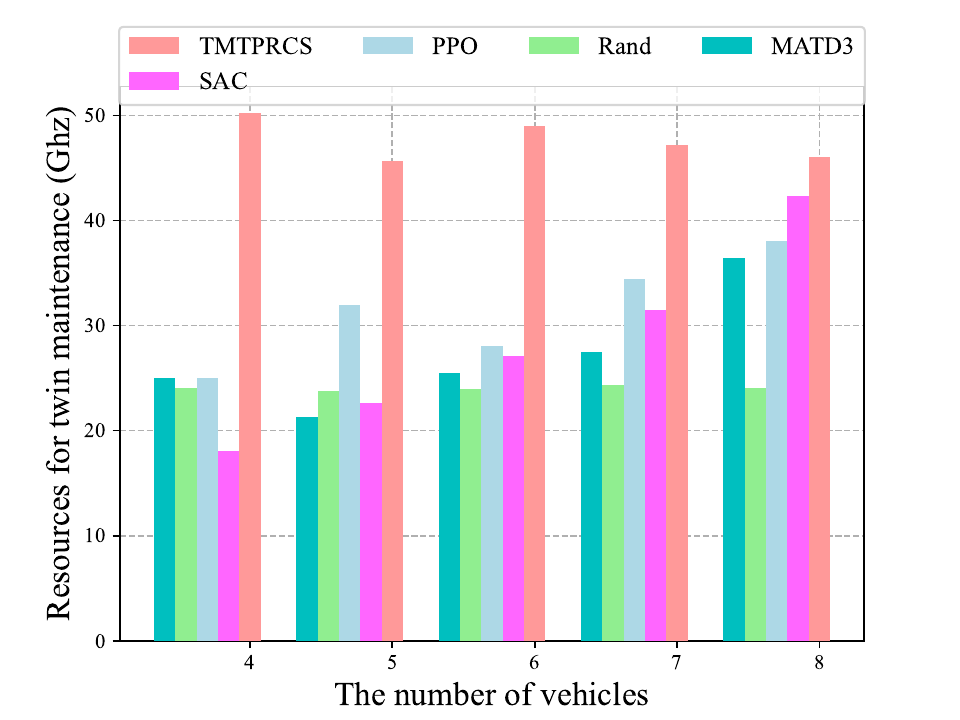}
				\label{figure5.b}}
		\end{subfigure}
		\centering
		\begin{subfigure}[Schedulable resources for tasks processing]
			{
				\centering
				\includegraphics[width=3.0in]{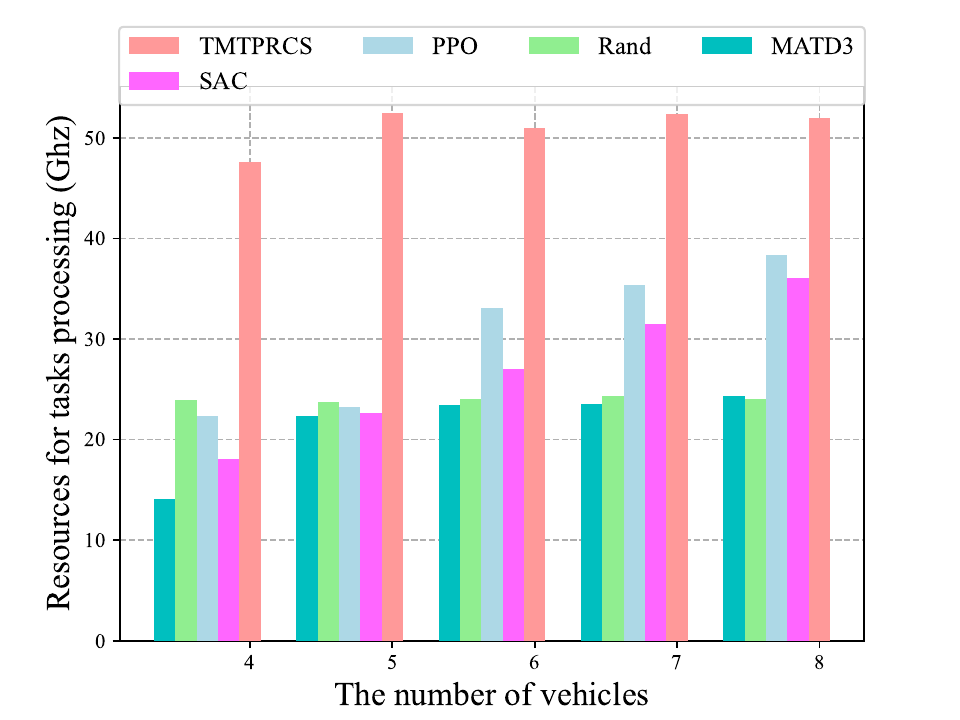}
				\label{figure5.c}}
		\end{subfigure}
		\caption{Comparison of resource utilization and schedulable resources for twin maintenance/task processing under different algorithms}
		\label{figure5}
	\end{figure*}
	
	\begin{figure}[h]
		\centering
		\includegraphics[width=3.0in]{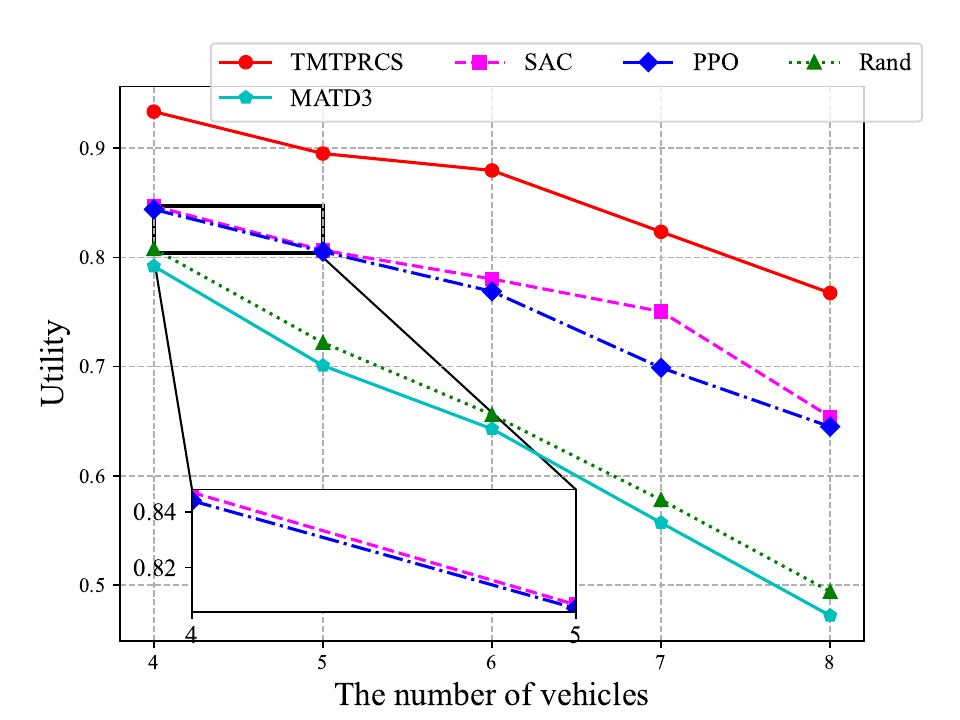}
		\caption{Utility under different algorithms}
		\label{figure6}
	\end{figure}
	
	\subsection{Performance Evaluation}
	Figure 4 compares the average rewards of the different algorithms. As the number of iterations gradually increases, the other three algorithms, except for the random algorithm, are able to increase from small to large values. This shows the effectiveness of DRL. In addition, TMTPRCS achieves higher reward values than SAC, PPO and Random algorithms. Compared to SAC, PPO, Random and MATD3 algorithms, the reward value achieved by TMTPRCS after convergence is $6.42\%$, $11.15\%$, $37.56\%$ and $83.42\%$ higher than them respectively. In a fully competitive setting, our method works well to obtain the optimal policy whereas MATD3 converges to a locally optimal solution due to insufficient exploration.
	
	Figure 5 compares the resource utilization of different algorithms for different number of vehicles. As can be seen in Fig. 5.a, TMTPRCS always maintains a high resource utilization compared to SAC ,PPO, and MATD3. On the contrary, SAC, PPO and MATD3 only gradually increase its resource utilization when the number of vehicles gradually increases. For TMTPRCS, the reason that the resource utilization is higher than the other numbers when the number of vehicles is 6 is that the algorithm achieves the optimal resource allocation balance in this case, i.e., there are neither idle nor overloaded resources. Specifically, when the number of vehicles is lower than 6, the number of twin models to be maintained in the server is lower, which results in less resource demand and, therefore, slightly lower utilization of resources. When the number of vehicles is higher than 6, the demand for resources increases, but the total resources in the server are limited, thus leading to a decrease in the overall utilization of resources. This numerical result also indicates that when the resources of twin model are limited, there exists an optimal served number of vehicles, which needs to be determined in theory in the future. In fact, TMTPRCS is able to collect global information about all vehicles and the requirements corresponding to twin maintenance tasks and computation tasks due to the centralized training mentioned in Fig.4. On the contrary, PPO and SAC only make decisions based on the local information of a single vehicle and share it directly to other vehicles. In addition, MATD3, as described in Fig. 4, it converges to a locally optimal solution due to under-exploration, thus leading to under-utilization of resources under this strategy. In order to visualize the resources that can be dispatched by the vehicles when performing twin maintenance and task processing, we present Fig. 5.b and Fig. 5.c. It can be seen that the amount of resources that can be scheduled by TMTPRCS is higher than that of the other three.
	
	Figure 6 illustrates the maximum efficiency that can be achieved for each vehicle at different numbers of vehicles. As the number of vehicles gradually increases, the effectiveness achieved per vehicle decreases. Obviously, an increase in the number of vehicles implies an increase in the demand for twin maintenance and task processing. Due to the resource competition between vehicles, vehicles will always request more resources to avoid the failure of these two tasks. However, the total resources of the server remain the same, so under different algorithm scheduling, the resources that each vehicle can request will also decrease, which in turn decreases the performance. In addition, TMTPRCS can better cope with competitive situations with multiple vehicles because it can be based on global information.
	
	\begin{figure}[!t]
		\centering
		\begin{subfigure}[Delay of twin maintenance ]{
				\centering
				\includegraphics[width=3.0in]{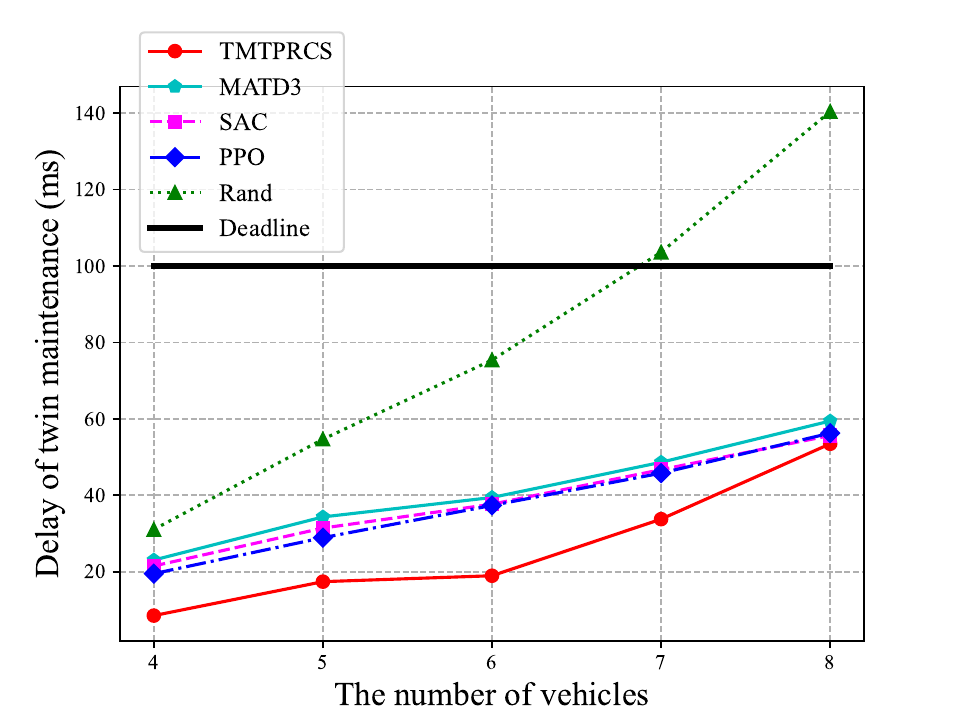}
				\label{figure7.a}}
		\end{subfigure}
		\centering
		\begin{subfigure}[Delay of vehicular computing task processing]
			{
				\centering
				\includegraphics[width=3.0in]{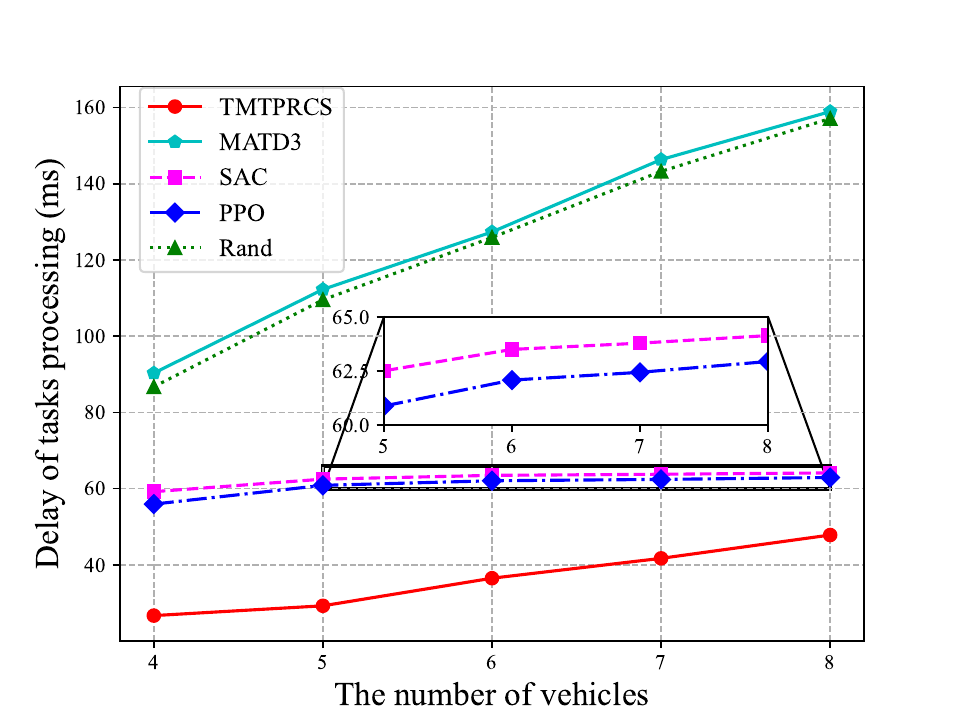}
				\label{figure7.b}}
		\end{subfigure}
		\caption{Comparison of two types of delays under different algorithms:$p_i$=200mW}
		\label{figure7}
	\end{figure}
	
	Figure 7 compares the twin maintenance delay and task processing delay of different algorithms for different number of vehicles. It can be seen that both types of delays gradually increase with the number of vehicles. Specifically, in Fig. 7.a, all four types of algorithms, except for the random algorithm, meet the deadline for the twin maintenance task. Here, the reason for the slower delay growth when the number of vehicles increases from 5 to 6 is that the algorithm is able to allocate resources more efficiently in this case, reducing resource competition. Specifically, the delay growth rate in this range is only $8.91\%$, significantly lower than the previous stage. In contrast, when the number of vehicles increases from 4 to 5, the demand for resources rises while the allocation is not yet optimal, resulting in a high delay growth rate of $104.41\%$, although the total delay remains lower than that observed at 5 and 6 vehicles. Similarly, as the number of vehicles increases from 6 to 7, the resource demand continues to rise, and competition intensifies, leading to a significant $78.16\%$ increase in delay. This trend highlights the limits of the algorithm's resource management under heavier loads. Nevertheless, despite these variations in delay growth, the overall latency remains within acceptable bounds. And in Fig. 7.b, all five classes of algorithms satisfy the maximum time tolerance for the computation task. In addition, TMTPRCS always achieves lower latency compared to the other three classes of algorithms. This is due to the fact that on the one hand, according to Fig. 5, TMTPRCS has a higher resource utilization and thus more schedulable resources, and on the other hand, according to what is described in Fig. 6, TMTPRCS is able to better deal with resource competition under multiple vehicles. From the two figures, on the one hand, it can be seen that the MATD3 algorithm has the highest task processing latency due to the low resource utilization of the MATD3 algorithm and on the other hand, the locally optimal solution achieved by MATD3 has made it invest its resources in reducing the twin maintenance delay.	
	
	\section{Conclusions}
	This paper considered the server resource allocation problem of twin maintenance and computing task processing simultaneously under edge computing for a digital twin scenario with single server and multiple vehicles, and constructed it as an optimization problem to maximize the resource utility of each vehicle. In the future, we will conduct a more in-depth study. In multi-server scenarios, servers can share resource and load data to collaborate. Dynamic load balancing can help assign tasks in real time. In this setting, When TMTPRCS shifts from centralized to decentralized training, it brings partial observability challenges to the critic network. These can be mitigated through joint updates, global state aggregation, or hierarchical coordination. In practical deployments, as vehicles move across base stations, their digital twin models need real-time migration. We aim to enhance TMTPRCS with trajectory-aware resource pre-allocation, lightweight state transfer, and decentralized edge server coordination to support mobility. In addition, while TMTPRCS currently assumes offline centralized training, future work will focus on adapting the algorithm for real-world edge deployments. This includes exploring model compression techniques, lightweight network architectures, and shared policy strategies to reduce computational and memory overhead on edge servers. At the same time, comparisons with classical reinforcement learning methods are insufficient, so we will introduce comparisons with more advanced methods in future work.
	
	The conclusion is summarized as follows:
	\begin{itemize}
		\item Our method can make real-time decisions on moving vehicles and make decisions on multiple vehicles in the environment simultaneously.
		\item Our approach enables the development of rational resource request policies for each vehicle under twin maintenance and vehicular tasks deadlines and total resource constraints.
	\end{itemize}

\end{document}